\documentclass{article}
\usepackage{spconf,amsmath,graphicx}
\usepackage{booktabs}
\usepackage{cite}
\usepackage{xcolor}
\usepackage{algorithmic}
\usepackage{graphicx}
\usepackage[ruled,vlined]{algorithm2e}   
\usepackage{amssymb}
\usepackage{mathtools}
\usepackage{caption}
\usepackage{hyperref}
\usepackage{cleveref}
\newtheorem{theorem}{Theorem}[section]

\newtheorem{lemma}[theorem]{Lemma}

\newtheorem{definition}[theorem]{Definition}
\newtheorem{assumption}[theorem]{Assumption}
\newtheorem{fact}[theorem]{Fact}

\newcommand\bLambda{\boldsymbol{\Lambda}}
\newcommand\bzeta{\boldsymbol{\zeta}}
\newcommand\tldx{\widetilde{\mathbf{x}}}

\newcommand\mcf{\mathcal{F}}

\newcommand\bz{\mathbf{z}}

\newcommand\bE{\mathbb{E}}

\newcommand\bx{\mathbf{x}}
\newcommand\bw{\mathbf{w}}
\newcommand\by{\mathbf{y}}

\newcommand\bI{\mathbf{I}}

\newcommand\prox{\mathrm{prox}}

\definecolor{purple}{rgb}{0.6, 0.2, 0.8}
 
\usepackage{multibib}
\newcites{main}{refs}
\newcites{appendix}{refs}
\title{Dynamic Privacy Allocation for Locally Differentially Private \\Federated Learning with Composite Objectives}
\name{Jiaojiao Zhang, Dominik Fay and Mikael Johansson \thanks{J. Zhang, Dominik Fay, and M. Johansson are with the Division of Decision and Control
		Systems, School of Electrical Engineering and Computer Science, KTH
		Royal Institute of Technology, SE-100 44 Stockholm, Sweden. Email:
		\{jiaoz,dominikf,mikaelj\}@kth.se. This work was supported in part by the funding from Digital Futures, WASP, and  VR under the contract 2019-05319. }}
\address{School of Electrical Engineering and Computer Science, KTH
	Royal Institute of Technology }
\begin{document}
%
\maketitle
\begin{abstract}
This paper proposes a locally differentially private 
federated learning algorithm for strongly convex but possibly nonsmooth problems that protects the gradients of each worker against an honest but curious server. The proposed algorithm adds artificial noise to the shared information to ensure privacy and dynamically allocates the time-varying noise variance to minimize an upper bound of the optimization error subject to a predefined privacy budget constraint. This allows for an arbitrarily large but finite number of iterations to achieve both privacy protection and utility up to a neighborhood of the optimal solution, removing the need for tuning the number of iterations. Numerical results show the superiority of the proposed algorithm over state-of-the-art methods.
\end{abstract}
\section{Introduction}
Federated Learning (FL) \cite{li2020federated} is a popular distributed machine learning framework where a set of workers collaborate on model training through a server without sharing their local data. Not only does this enable efficient parallel computations, but it also provides privacy advantages compared to centralized machine learning methods.  
Unfortunately, even though the data is not directly shared, FL can still result in privacy leakage \cite{zhu2019deep}. This is because the information communicated between the workers and the server,  such as local gradients or models, is derived from the raw data. As the number of iterations increases, the accumulated shared information during the training process may pose a privacy risk \cite{zhu2019deep}. 

One way to approach privacy protection is through the lens of cryptography. However, in many FL scenarios, these techniques suffer from  
high communication and computation costs \cite{chou2020privacy}. 
Currently, Differential Privacy (DP) \cite{dwork2006calibrating,dwork2008differential} has been more successful in FL, since it is easy to implement, has a good privacy protection effect, and allows to rigorously quantify the strength of the privacy protection \cite{dwork2008differential}.  
The design of DP algorithms in federated learning depends on the attack     scenario and can be roughly divided into global DP and local DP (LDP) \cite{hu2021concentrated, zhao2022fldp}. Global DP resists passive attackers from outside the system and typically relies on the server to add noise to the aggregated information while the workers upload their true models or gradients, assuming that the upload communication channel is secure and the server is trustworthy \cite{hu2021concentrated}. In contrast, LDP protects sensitive information against an honest but curious server during the training process and does not rely on secure uploading channels
\cite{zhao2022fldp}. Instead, each worker adds noise to its local sensitive information before sending it to the server. The challenge of LDP is that it may require more noise to be added, which can lead to a serious deterioration of the utility (e.g.~the expected value of the total loss function). A number of studies have been dedicated to proposing algorithms with better privacy-utility trade-off \cite{ 
 noble2022differentially,lowy2022private}. 
One example of such a technique is the recently proposed DP-SCAFFOLD~\cite{noble2022differentially}, which combines LDP with SCAFFOLD \cite{karimireddy2020scaffold}. It analyzes the trade-off between privacy and utility and demonstrates the superior performance of DP-SCAFFOLD compared to DP-FedAvg \cite{mcmahan2018learning}, especially for heterogeneous data. The work in \cite{lowy2022private} proposes ISRL-DP to solve problems with convex constraints.  
However, all the works mentioned above face the issue that  the optimization error bound is sensitive to the total number of iterations when the privacy budget is finite.
Therefore, the total number of iterations becomes a hyperparameter, which needs to be  chosen carefully to minimize the optimization error. Either too few or too many iterations reduce the utility, which brings difficulties to the practical use of the algorithm.  
\begin{table*}[htbp]
	\begin{center}
		\caption{Comparison with state-of-the-art methods satisfying $(\epsilon ,\delta)$-LDP.}
		\label{tab-1}
		\setlength{\tabcolsep}{4pt}
		\begin{tabular}{lcccr}
			\toprule
			\bf Algorithm  &	\bf Choice of $T$ & \bf Utility\footnotemark[1]
			& \bf Composite Problems \\ 
			\midrule 			 		
			DP-FedAvg\cite{noble2022differentially} & requires a certain $T^{\star}$  & $\mathcal{O}\left(\hat{C}^2 \frac{1 }{\mu_f} \frac{d\ln(1/\delta)}{\epsilon^2 n m^2}\right)$  & no  \\ 
			DP-SCAFFOLD \cite{noble2022differentially}& requires a certain $T^{\star}$ & $\mathcal{O}\left(\hat{C}^2 \frac{1 }{\mu_f} \frac{d\ln(1/\delta)}{\epsilon^2 n m^2} \right)$  & no  \\
			ISRL-DP \cite{lowy2022private}	& requires a certain $T^{\star}$ & $\mathcal{O}\left({\hat{L}^2 }\frac{1 }{\mu_f} \frac{d\ln(1/\delta)}{\epsilon^2 n m^2} \right)$  & no \\
			Proposed & allows all $T\ge  \tfrac{\ln \left(\tfrac{\epsilon^2nm^2\min\{{\mu_f^2},1\} \bE[ \Phi^{1} ]}{B^2 d \ln ({1}/{\delta})} \right)}{\gamma \min\{\mu_f,1\}}$ & $\mathcal{O}\left(B^2 \frac{1}{\min\left\{{\mu_f^2},1\right\}}  \frac{d \ln(1/\delta)}{\epsilon^2  n{m^2}}\right)$ & yes \\
			\bottomrule
		\end{tabular}
	\end{center}
\end{table*} 
\footnotetext[1]{$\hat{C}$ is the gradient clipping threshold; $T$ is the rounds of communication; $\hat{L}$ is the Lipschitz continuous constant of the local loss functions.}

{\bf Contribution.}  We propose an LDP FL algorithm for strongly convex but possibly nonsmooth problems that protects worker gradients from an honest but curious server. 
Our algorithm adds artificial noise to shared information for privacy protection. A dynamic allocation of the noise variance minimizes the optimization error under a predefined privacy budget constraint.
As a result, the optimization error caused by the privacy-protecting noise is independent of the total number of iterations, allowing us to execute an arbitrarily large but finite number of iterations for privacy and utility without worrying that a too large number of iterations will cause the optimization error to diverge.  Our numerical results show a distinct superiority of the proposed algorithm over state-of-the-art methods.  Table \ref{tab-1} summarizes the comparison with the state-of-the-art methods. 

\textbf{Notation.}
We use the notation $\|\cdot\|$ for the Euclidean vector norm and $\otimes$ for the Kronecker product. Let $d$ and $n$ be positive integers. $I_d$ represents the $d\times d$ identity matrix, and $1_n$ (or $0_n$) represents the all-one (or all-zero) $n$-dimensional column vector. The set $[n]$ denotes the integers from $1$ to $n$.
The bold vector $\mathbf{x}$ aggregates the vectors $x_1,\ldots,x_n \in \mathbb{R}^d$ and is defined as $\mathbf{x}=[x_1;\ldots;x_n]\in \mathbb{R}^{nd}$. The subgradient of a function $g$ is denoted by $\partial g$. The expected value of a random variable $v$ is represented by $\bE [v]$, while $\bE [v|\mathcal{F}]$ denotes the conditional expectation of $v$ with respect to the event $\mathcal{F}$. The probability of a random event is denoted by $\operatorname{Pr}(\cdot)$, and $\mathcal{N}(\mu,\Sigma)$ represents a multivariate Gaussian distribution with mean $\mu$ and covariance matrix $\Sigma$. For two vectors $x$ and $y$, $x\le y$ means that each entry in $x$ is smaller or equal to the corresponding entry in $y$.  
\section{Problem Formulation and LDP}
In this section, we introduce the problem formulation for FL and review some preliminaries on LDP.
\vspace{-2mm}
\subsection{Problem Formulation}
In our FL scenario, a server coordinates $n$ workers to collaboratively train a joint model with $d$ parameters. Each worker, denoted by $i \in [n]$, possesses a private data set 
$\mathcal{D}_i = \bigcup_{l=1}^{m} \mathcal{D}_{il}$ comprising the $m$ data points  $\mathcal{D}_{i1}, \dots, \mathcal{D}_{im}$. We support heterogeneous data and do not require that the data sets  $\mathcal{D}_i$ are similar. Our goal is to solve composite problems on the form
\vspace{-1mm}
\begin{align}\label{eqn:basic_opt}
	\operatorname*{minimize}_{x\in \mathbb{R}^d} ~ \frac{1}{n}\sum_{i=1}^n f_i(x) + g(x), ~ f_i(x)=\frac{1}{m} \sum_{l=1}^{m} f_{il} (x).
\end{align}
In this context, $f_{il}(x)$ represents the sample loss associated with data point $\mathcal{D}_{il}$, while the loss function $f_i(x)$ measures the dissimilarity between the model's predictions using parameter $x$ and the actual labels in the data set $\mathcal{D}_i$. Additionally, $g$ serves as a regularizer that promotes desired properties of the solution, such as sparsity.
To address \eqref{eqn:basic_opt} in the FL setting, we introduce a local copy $x_i$ of the decision vector $x$ for each worker $i$ and reformulate the problem as
\begin{equation}\label{eqn:opt_consensus}
	\begin{aligned}	
		\operatorname*{minimize}_{\{x_1,\ldots,  x_{n}\}}~ & \frac{1}{n}\sum_{i=1}^n \left(f_i(x_i)+g(x_i)\right),\\
		\text{s.t.}\hspace{5mm}  & x_i-\bar{x}=0, ~\forall i\in[n], 
	\end{aligned} 
\end{equation}
where $\bar{x}=\sum_{i=1}^{n} x_i/n$.
To simplify the process of developing and analyzing the algorithm, we rewrite \eqref{eqn:opt_consensus} in a compact form. We let $\tilde{L}=I_n-1_n 1_n^T/n$ and represent the local copies $x_i\in {\mathbb R}^d$ as an aggregated vector $\bx \in \mathbb{R}^{nd}$. By defining
\begin{align*}
 f(\bx) = \frac{1}{n} \sum_{i=1}^n f_i(x_i),~
	g(\bx) = \frac{1}{n}\sum_{i=1}^n  g(x_i), ~\mathbf{L}=\tilde{L}\otimes I_d,
\end{align*}
we can express (\ref{eqn:opt_consensus}) in the form 
\begin{equation}\label{eqn:opt_consensus_compact}
	\begin{aligned}
		\operatorname*{minimize}_{\bx\in \mathbb{R}^{nd}}~  f(\bx)+g(\bx), 
		~\mbox{s.t.} ~ \mathbf{L}\bx=0_{nd}.
	\end{aligned}
\end{equation}
By the definition of $\mathbf{L}$, the constraint $\mathbf{L}\bx=0_{nd}$ is equivalent to $x_1=\cdots=x_n=\bar{x}$. 
In this reformulation \eqref{eqn:opt_consensus_compact}, each worker has to minimize its local loss (defined by $f_i(x_i)$ and $g(x_i)$) and cooperates with the others to reach consensus on the globally optimal decision variable.  
Throughout the paper, we make the following assumptions on $f$ and $g$. 
\begin{assumption}\label{asm-convex}
	The loss function $f: \mathbb{R}^{nd} \mapsto \mathbb{R}$ is both $\mu_f$-strongly convex and $L_f$-smooth, \emph{i.e.} 
	for any $\bx,\by\in \mathbb{R}^{nd}$, 
$$f(\by)\geq f(\bx) + \nabla f(\bx)^T(\by-\bx) + \frac{\mu_f}{2}\|\by-\bx\|^2,
$$
\vspace{-2mm}
$$
f(\by)\leq f(\bx) + \nabla f(\bx)^T(\by-\bx) + \frac{L_f}{2}\|\by-\bx\|^2,
$$
	where $\mu_f>0$ and $L_f>0$ denote the strong convexity constant and smoothness constant, respectively.     Additionally, we assume that $g: \mathbb{R}^d \mapsto \mathbb{R}$ is a closed and proper convex function, which may not necessarily be smooth.
\end{assumption}

\subsection{Local Differential Privacy}
We consider a threat model in which an honest-but-curious server has access to all the messages shared by the workers during the training process, as studied in \cite{noble2022differentially}.
%
We aim to protect each gradient $\nabla f_{il}$ for all $i\in[n]$ and $l\in [m]$. Thus, 
we say that $\nabla f_i$ and $\nabla f_i'$ is a pair of adjacent inputs of worker $i$ if only a single gradient (e.g. $\tilde{l}$-th) is different, i.e., $\nabla f_{i\tilde{l}}\neq \nabla f'_{i\tilde{l}}$ while $\nabla f_{il}= \nabla f'_{il}$ for any $l\neq \tilde{l} $.  
With this definition of adjacent inputs, we give the following definition of LDP.  
\begin{definition}[$(\epsilon, \delta)$-LDP]
	\label{def-dp} 
	For any adjacent inputs $\nabla f_i$ and $\nabla f_i'$ of worker $i$, a randomized algorithm $\mathcal{A}$ satisfies $(\epsilon, \delta)$-LDP if for 
	every $O \subset  \text{Range}(\mathcal{A})$ where $\text{Range}(\mathcal{A})$ is the output space of $\mathcal{A}$,  we have 
	\begin{equation}\label{eq-zo}
		{\operatorname{Pr}[\mathcal{A}(\nabla f_i)\in O]} \le e^\epsilon {\operatorname{Pr}[\mathcal{A}(\nabla f_i')\in O]}+\delta. 
	\end{equation}
\end{definition}
When $\epsilon$ and $\delta$ are sufficiently small, the outputs generated by any pair of adjacent inputs $\nabla f_i$ and $\nabla f_i'$  are almost identical, which makes it difficult for attackers to distinguish between them. One prototypical method for achieving $(\epsilon, \delta)$-LDP is the Gaussian Mechanism, which adds Gaussian noise $ \mathcal{N}(0, \xi^2 I_d)$ to a real-valued query function $\mathcal{Q}: \nabla f_i \mapsto \mathbb{R}^d$ to perturb the true answer. The variance of the noise that one needs to add depends on the sensitivity of $\mathcal{Q}$ \cite{ dwork2006calibrating}. 
\begin{definition} [Sensitivity] 
	\label{def-sensitivity}	
	For any adjacent inputs $\nabla f_i$ and $\nabla f_i'$, the sensitivity of a query function $\mathcal{Q}$ is defined as
	$	\Delta_{\mathcal{Q}}=\max_{\nabla f_i, \nabla f_i'}\|\mathcal{Q}(\nabla f_i )-\mathcal{Q}(\nabla f_i')\|.
	$
\end{definition}
To ensure a desired level of privacy, a higher variance is needed when the sensitivity $\Delta_{\mathcal{Q}}$ is higher \cite{dwork2006calibrating}. 
Applying a differentially private mechanism repeatedly increases the privacy loss. We use zero-concentrated differential privacy ($\rho$-zCDP \cite{bun2016concentrated}) as an analytical tool to track the privacy loss of our algorithm across iterations. The $\rho$-zCDP captures the privacy properties of repeated Gaussian mechanisms better than analyzing each iteration separately via $(\epsilon, \delta)$-LDP \cite{bun2016concentrated}. The total $\rho$-zCDP bound can then be converted back to $(\epsilon, \delta)$-LDP.

\section{Algorithm Development}
\begin{algorithm}[htbp]
	\caption{The proposed algorithm }
	\label{alg-fl}
	\begin{algorithmic}
		\STATE $\textbf{Input:}$ $\gamma$,  $x_i^0$, $\Lambda_i^0=0_d$; $\xi^2_t$ follows dynamic allocation.
		\FOR {$t = 1, 2, \ldots, T$ }
		\STATE {\bf Worker $i$}
		\STATE Generate the noise $\zeta_i^t \sim \mathcal{N}\left(0, \xi_t^2 I_d\right)$
		\STATE Update $\widetilde{x}_i^{t+1} =x_i^t-\gamma\left(\frac{1}{n}\nabla f_i(x_i^t)+\zeta_i^t+\Lambda_i^t\right)$
		\STATE Send $ \widetilde{x}_i^{t+1}$ to the server
		\STATE {\bf Server}
		\STATE Compute  $\overline{\widetilde{x}}^{t+1}=\frac{1}{n} \sum_{i=1}^{n}\widetilde{x}_i^{t+1}$ and broadcast $\overline{\widetilde{x}}^{t+1}$  to all the workers	
		\STATE {\bf Worker $i$}
		\STATE Receive $\overline{\widetilde{x}}^{t+1}$ from the server
		\STATE Update $\Lambda_i^{t+1}=\Lambda_i^t+\widetilde{x}_i^{t+1}- \overline{\widetilde{x}}^{t+1} $
		\STATE Update $z_i^{t+1} =\widetilde{x}_i^{t+1}-\gamma\left(\widetilde{x}_i^{t+1}  -\overline{\widetilde{x}}^{t+1}\right) $
		\STATE  Update $x_i^{t+1}=\prox_{\frac{\gamma}{n}g}(z_i^{t+1})$
		\ENDFOR
		\STATE \textbf{Output:} $x_i^{T+1}$
	\end{algorithmic}
\end{algorithm}
Next, we introduce our LDP FL algorithm for solving \eqref{eqn:basic_opt} with a predefined privacy budget.
The algorithm relies on the following primal-dual formulation of~\eqref{eqn:opt_consensus_compact}
 \vspace{-1mm}
\begin{equation}\label{eq-saddle-point}
	\operatorname*{minimize}_{\bx\in \mathbb{R}^{nd}}\;  \operatorname*{maximize}_{\by \in \mathbb{R}^{nd}} ~ f(\bx)+g(\bx)+\langle \by, \mathbf{L} \bx\rangle, 
\end{equation} 
where $\by \in \mathbb{R}^{nd}$ is the dual variable of the equality constraint in~(\ref{eqn:opt_consensus_compact}). 
To solve \eqref{eq-saddle-point}, we adopt an approach that alternates between computing an approximate primal solution for a fixed dual variable and performing a (sub)gradient ascent step in the dual variable for a fixed primal variable~\cite{alghunaim2019linearly,mishchenko2022proxskip}. In addition, we use the Gaussian mechanism to protect gradients. In principle, we would like to run the iterations
 \vspace{-1mm}
\begin{equation}\label{eq-alg-fl}
	\hspace{-1mm}\left\{ 
	\begin{aligned}
		\tldx^{t+1}&=\bx^t-\gamma\left(\nabla f(\bx^t)+\bzeta^t +\mathbf{L}\by^t  \right),\\
		\by^{t+1}&=\by^t+\mathbf{L} \tldx^{t+1},\\
		\bz^{t+1}&=\left(\bI-\gamma \mathbf{L}\right)\widetilde{\bx}^{t+1},\\
		\bx^{t+1}&=\prox_{\frac{\gamma}{n}g}( \bz^{t+1}), 
	\end{aligned}
	\right.
\end{equation}
where 
$\nabla f(\bx^t)=[\frac{1}{n}\nabla f_1(x_1^t);\ldots;\frac{1}{n}\nabla f_n(x_n^t) ] \in \mathbb{R}^{nd}$, $\bzeta^t = [\zeta_1^t;\ldots; \zeta_n^t]\in \mathbb{R}^{nd}$, and $\zeta_i^t \sim \mathcal{N}\left(0, \xi_t^2 I_d\right)$ for all $i\in [n]$. Note that the proximal operator, $\prox_{\frac{\gamma}{n}g}(z_i)=\arg\min_{u\in \mathbb{R}^d} \frac{\gamma}{n} g(u)+\frac{1}{2}\|z_i-u \|^2 $, in \eqref{eq-alg-fl} 
is defined block-wise so that $x_i^{t+1}=\prox_{\frac{\gamma}{n}g}(z_i^{t+1})$ for all $i\in[n]$. 

In the compact formulation (\ref{eq-alg-fl}), any multiplication with $\mathbf{L}$ necessitates communication with the server. To ensure LDP, we would therefore need to also protect the dual variables $\by^{t}$ when they are communicated to the server to evaluate $\mathbf{L}\by^{t}$. However, since $\mathbf{L}^2=\mathbf{L}$, we can introduce  $\bLambda^t=\mathbf{L}\by^t$
with $\bLambda^1=0_{nd}$ and re-write the iterations \eqref{eq-alg-fl}  as
 \vspace{-1mm}
\begin{equation}\label{eq-alg-Lambda}
	\left\{ 
	\begin{aligned}
		\tldx^{t+1}&=\bx^t-\gamma\left(\nabla f(\bx^t)+{\bzeta^t}+ \bLambda^t\right),\\
		\bLambda^{t+1}&=\bLambda^t+\mathbf{L} \tldx^{t+1},\\
		\bz^{t+1}&=\left(\bI-\gamma \mathbf{L}\right)\widetilde{\bx}^{t+1},\\
		\bx^{t+1}&=\prox_{\frac{\gamma}{n}g}( \bz^{t+1}).
	\end{aligned}
	\right.
\end{equation}
Unlike \eqref{eq-alg-fl}, the updates in \eqref{eq-alg-Lambda} only require the communication of the  obfuscated local models $\widetilde{\bx}^{t+1}$ to  update  $\bLambda^{t+1}$ and $\bz^{t+1}$. The implementation of \eqref{eq-alg-Lambda} is given in Algorithm \ref{alg-fl}.  

 \vspace{-1mm}
\section{ Analysis} \label{sec:analysis}
In this section, we first perform separate analyses of the privacy and utility of the proposed algorithm. Then, we perform a joint privacy-utility analysis under a dynamic allocation of the variance of the injected noise and give the resulting convergence rate.  {All proofs can be found in Appendix \ref{section-app}. }

\subsection{Separate Analysis of Privacy and Utility} 
We make the following assumption on the injected noise.
\begin{assumption}\label{asm-variance}
	At any time step $t$ and for every worker $i \in [n]$, the injected noise variables $\zeta_i^t$ are independent and follow a Gaussian distribution with mean 0 and variance $\xi_t^2 I_d$. 
\end{assumption}
To ensure bounded sensitivity, we need $\nabla f_i(x_i^t)$ to be bounded for all $i$ and $t$. The convexity of $f_i$ simplifies this requirement to the boundedness of $x_i^t$ \cite[Prop. B.24]{Bertsekas2003}. To achieve this, we make specific assumptions on the regularizer $g$.
	\begin{assumption}\label{asm:g-prox}
  The proximal operator of the regularizer,  $\prox_g(z)=\arg\min_{x\in \mathbb{R}^d} g(x)+\tfrac{1}{2}\|z-x \|^2 $,
  is both easy to compute and bounded.
	\end{assumption}
We present two examples of $g$ that satisfy Assumption \ref{asm:g-prox}:

\noindent
(1) The indicator function of a convex and compact set which is easy to project onto. When $g$ is the indicator function of a convex set,  the proximal operator reduces to the projection operator, which is bounded if the underlying set is.

\noindent(2) The weighted $\ell_1$-norm over a box \cite[Example 6.23]{beck2017first})  
 \vspace{-1mm}
	$$
	g(z)= \begin{cases}\sum_{j=1}^d \omega_j\left|z_j\right|, & -{\alpha}1_d \leq z \leq {\alpha}1_d, \\ \infty, & \text { else, }\end{cases}
	$$
	where $\omega_j >0$, $\alpha >0$, and $z_j\in \mathbb{R}$ is the $j$-th entry of $z$. We can express $g(z)$ as  $g(z)=\sum_{j=1}^d g_j(z_j)$, where
	$
	g_j(z_j)= \omega_j|z_j|$ if $  -\alpha \leq z_j \leq \alpha, 
	$  and $g_j(z_j)=\infty$ otherwise.
	The $j$-th entry of $\prox_g(z) $, defined by $[\prox_g(z)]_j$, is given by
	$$[\prox_g(z)]_j=\min \left\{\max \left\{\left|z_j\right|-\omega_j, 0\right\}, \alpha\right\} \operatorname{sign}\left(z_j\right),$$
 Consequently, we have $ -\alpha \le [\prox_g(z)]_j \le \alpha$, and the proximal operator of $g$ is bounded and easy to evaluate.

With Assumption \ref{asm:g-prox}, we know that there exists a constant $0<B<\infty$ so that, for all possible sequences $\{ \bx^t\}_t$ generated by Algorithm~\ref{alg-fl}, 
 \vspace{-1mm}
\begin{equation}
	\| \nabla f_{il}(x_i^t)\|\le B~ \mbox{ for all } i,t \mbox{ and } l.
\end{equation}
Since we aim to ensure that the messages $\{ \widetilde{x}_i^{t} \}_t$ preserve $(\epsilon,\delta)$-LDP for each gradient $\nabla f_i$, we establish the following lemma to bound the total privacy loss. 
\begin{lemma}\label{lem-total-dp}
	Let Assumptions \ref{asm-variance} and \ref{asm:g-prox} be satisfied.
	Then, for each worker $i$, the message sequence $\{ \widetilde{x}_i^{t} \}_t$ generated by Algorithm \ref{alg-fl}
	is $(\epsilon,\delta)$-LDP
	for any $\delta>0$, 
	with $\epsilon=\rho_{\rm tgt}+2 \sqrt{\rho_{\rm tgt} \ln ({1}/{\delta})}, $
	where $\rho_{\rm tgt}=\sum_{t=1}^T
	2B^2/(n^2m^2\xi_t^2)$.
\end{lemma}
Although we state this guarantee for the messages $\{ \widetilde{x}_i^{t} \}_t$, it also holds for the iterates $\{ x_i^t\}_t$ because $x_i^t$ is a data-independent post-processing of $\widetilde{x}_i^{t}$ \cite{dwork2008differential}.
As shown in Lemma \ref{lem-total-dp}, the larger the variance $\xi_t^2$, the stronger the privacy protection.  Further, we have the freedom to allocate the variances $\xi_t^2$ over time to guarantee that $\rho_{\rm tgt}$, and hence the total privacy loss $\epsilon$, is no more than a given privacy budget.

Next, we focus on the utility. To handle the randomness caused by $\zeta_i^t$ for all $i\in [n]$, we let $\mathcal{F}^t$ be the event generated
by $\bigcup_{i \in [n]}^{\tilde{t} \in [t-1]} \zeta_i^{\tilde{t}}$.  
Let $(\bx^{\star},\by^{\star})$ be a pair of primal-dual optimal solutions to the constrained problem  \eqref{eqn:opt_consensus_compact}. 
We define the Lyapunov function $\Phi^{t}=   \left\|\bx^{t}-\bx^{\star}\right\|^{2}+{\gamma}\left\|\by^{t}-\by^{\star}\right\|^{2}$
and derive the following recursion. 
\begin{lemma}\label{lem-descent}
	Under Assumptions \ref{asm-convex} and \ref{asm-variance}, for Algorithm \ref{alg-fl}, if 
$		\gamma\le \min\left\{ {1}/{4}, {1}/{L_f}\right\},
$
	then the sequence $\{\Phi^t\}_t$ satisfies
	\begin{equation}\label{eq-one-descent}
		\hspace{-1mm}\begin{aligned}
			\bE\left[\Phi^{t+1}\right]
			\le\left( 1-\gamma \min\left\{{\mu_f},1\right\} \right) \bE\left[\Phi^t \right]+\tfrac{5}{2}{nd\gamma^2}\xi_t^2. 
		\end{aligned}
	\end{equation}	
\end{lemma}

Lemma \ref{lem-descent} demonstrates that $\bE\left[\Phi^{t+1}\right]$ 
decreases by a factor $\left( 1-\gamma \min\left\{{\mu_f},1\right\} \right)$ but is also inflated by an additive term
$\tfrac{5}{2}{nd\gamma^2}\xi_t^2$ caused by the privacy noise.   By Lemma \ref{lem-total-dp}, if we use a static variance, \emph{i.e.,} let  $\xi_t^2=2B^2 T/ (\rho_{\rm tgt}n^2 m^2),~ \forall t\in [T]$, to meet the target value of $\rho_{\rm tgt}$ over the $T$ iterations,  the error term in the optimization error bound caused by privacy-protecting noise will diverge as $T$ grows large, as shown in \cite{noble2022differentially,lowy2022private}. However, by telescoping \eqref{eq-one-descent}, we observe that noise added earlier has a smaller effect on the total error at iteration $T$. This observation inspires us to allocate larger variances at earlier stages. We develop the precise mechanism for variance selection next.  
\subsection{Privacy-Utility Analysis with Dynamic Allocation}
Lemma \ref{lem-total-dp} characterizes the amount of variance required to ensure $(\epsilon,\delta)$-LDP, while Lemma \ref{lem-descent} quantifies the impact of the noise variance on the utility. Next, we conduct a joint analysis of the privacy and the utility under a dynamic allocation of the time-varying variances.
\begin{theorem}\label{thm-final}
	Let Assumptions \ref{asm-convex}, \ref{asm-variance}, and \ref{asm:g-prox} be satisfied.
	Then, for an arbitrarily large but finite $T$ and given privacy parameters $\epsilon > 0$ and $\delta \in (0, 1)$,
	if 
 \vspace{-1mm}
	\begin{equation}\label{eq-gamma}
		\gamma\le \min\left\{ {1}/{4}, {1}/{L_f}\right\}
	\end{equation}
	and the noise variances are allocated as 
 \vspace{-1mm}
	\begin{equation}\label{eq-variance}
		\xi_t^2= \sqrt{{\pi}/{q_t} },
		\quad \forall t\in [T],
	\end{equation}
	where $q_t=(1-\gamma \min\{\mu_f,1\})^{T-t}$ and 
	$\sqrt{\pi}=$ $2B^2 \sum_{t=1}^T\sqrt{q_t}$ $(\rho_{\rm tgt} n^2m^2)^{-1}$, 
	then the utility of Algorithm \ref{alg-fl} satisfies
  \vspace{-1mm}
	\begin{equation}
		\begin{aligned}
			&\bE\left[ \Phi^{T+1} \right]\le (1-\gamma \min\left\{{\mu_f},1\right\}   )^T \bE\left[ \Phi^{1} \right]\\
			&+ \tfrac{20  B^2d }{ \left(\sqrt{\epsilon+\ln \tfrac{1}{\delta}}-\sqrt{\ln \tfrac{1}{\delta}}\right)^2 nm^2\min\left\{\mu_f^2,1\right\}}	
		\end{aligned}		
	\end{equation}
	and, for each worker $i$, the message sequence $\{ \widetilde{x}_i^{t} \}_t$ satisfies $(\epsilon,\delta)$-LDP
	with 
 \vspace{-2mm}	\begin{equation}\label{eq-epsilon}
		\epsilon=\rho_{\rm tgt}+2 \sqrt{\rho_{\rm tgt} \ln ({1}/{\delta})}.
	\end{equation}
\end{theorem}
With Theorem \ref{thm-final}, given an LDP privacy budget $(\epsilon, \delta)$, we first choose $\rho_{\rm tgt}$ such that \eqref{eq-epsilon} is satisfied. We then set $\gamma$ according to \eqref{eq-gamma}, and finally allocate the variances $\xi_t^2$ using \eqref{eq-variance}. With these choices, for any arbitrarily large but finite $T$, the sequence $\{\bx^t\}_t$ generated by Algorithm \ref{alg-fl} converges in expectation to a neighborhood of the optimal solution of \eqref{eqn:opt_consensus_compact} that is of order  $\mathcal{O}( {B^2 d \ln ( {1}/{\delta}})/(\epsilon^2nm^2\min\{{\mu_f^2},1\})) $, while also ensuring $(\epsilon, \delta)$-LDP. The dynamic allocation strategy used in the proposed algorithm employs a geometrically decreasing variance and allows to attain a stable performance also for large values of $T$; cf. Section~\ref{sec:smooth_problem} below.
\vspace{-2mm}
\section{Numerical Experiments}
We compare Algorithm \ref{alg-fl} with three state-of-the-art algorithms: DP-FedAvg \cite{noble2022differentially}, DP-SCAFFOLD   \cite{noble2022differentially}, and ISRL-DP\cite{lowy2022private}, which are all designed to resist an honest-but-curious server with a good privacy-utility trade-off but cannot handle nonsmooth terms such as $g$ in~\eqref{eqn:basic_opt}. Therefore, we conduct two sets of experiments. In the first set, we compare our algorithm with  DP-FedAvg, DP-SCAFFOLD, and ISRL-DP for smooth problems to demonstrate its  superiority. In the second set of experiments, we run our algorithm on nonsmooth problems under different settings to verify our theoretical results. We use standard benchmarking data sets from the libsvm repository\footnote{https://www.csie.ntu.edu.tw/\~{}cjlin/libsvmtools/datasets/}, run each experiment 20 times and plot the average results with 
error bars that represent the standard variance of the 20 runs.  The optimal solution $x^{\star}$ of problem \eqref{eqn:basic_opt} is calculated in advance. For all algorithms, we set the same privacy budget, $(\epsilon,\delta)$-LDP  with $\epsilon=1$ and $\delta=10^{-4}$, and compare their optimality errors defined as 
$$\text{optimality}:={\frac{1}{n} \sum_{i=1}^{n}{\|\bar{x}^t-x_i^t\|^2}+\frac{\|\bar{x}^t-x^{\star}\|^2 }{\|x^{\star}\|^2 }}.$$
This quantity is used to reflect both the consensus error and
the optimization error of the average.
For the compared methods, since each worker receives the global model $\bar{x}^t$ from  the server, the optimality error becomes ${\|\bar{x}^t-x^{\star}\|^2 }/{\|x^{\star}\|^2 }$.     

Additional  experiments on nonconvex nonsmooth problems can be found in Appendix \ref{section-app}.
\subsection{Smooth Problem} \label{sec:smooth_problem}
We consider an $\ell_2$-regularized logistic regression problem
\vspace{-2mm}
\begin{align}\label{eq-logreg}
\underset{{x} \in \mathbb{R}^{d}}{\operatorname{min}} \; \frac{1}{n}\sum_{i=1}^n f_i(x)+\frac{\vartheta_2}{2}\|x\|^2,
\vspace{-2mm}
\end{align}
where $f_i(x)=\tfrac{1}{m}\sum_{l=1}^{m} \ln \left(1+\exp \left(-\left(\mathbf{a}_{i l}^{T} {x}\right) b_{i l}\right)\right)$ is the loss of $x$ on the data held by worker $i$, $\vartheta_2$ is a regularization parameter, and $(\mathbf{a}_{i l}, b_{i l}) \in \mathbb{R}^{d} \times\{-1,+1\}$ is the feature-label pair of the $l$-th sample of worker $i$. 
We use the \texttt{ijcnn} dataset and set $n=20$, $m=100$, and $\vartheta_2=0.1$. 
We use full gradients in the learning process for all algorithms.  It is worth noting that subsampling amplification used in \cite{noble2022differentially,lowy2022private} helps to counteract the increased sensitivity that comes with using mini-batch stochastic gradients but it does not improve the expected utility beyond the use of full gradients.
We retain local updates for DP-FedAvg and DP-SCAFFOLD to reduce their communication costs. We set the privacy budget as $(1,10^{-4})$-LDP  and tune the parameters for the best performances. We evaluate 8  different values of the total communication rounds $T$, ranging from 1000 to 8000 with an interval of 1000. For each algorithm, we report the final optimization error after $T$ communication rounds. 
\vspace{-1mm}

\begin{figure}[htbp]
	\begin{minipage}[h]{1\linewidth}
	\centering
	\centerline{\includegraphics[width=5.6cm]{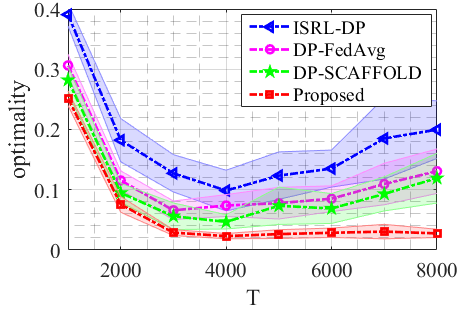}}
	\caption{Smooth problem on ijcnn satisfying $(1,10^{-4})$-LDP.}
\label{fig-T-8000}
\end{minipage}
\end{figure}

Fig.~\ref{fig-T-8000} shows the optimality error for  different numbers of total communication rounds $T$ for the considered methods. The state-of-the-art algorithms initially improve with more iterations, but then deteriorate if $T$ is chosen too large.  
In contrast,  the optimization error of our algorithm decreases  initially and stabilizes in a neighborhood of the optimum as $T$ increases, indicating that there is no need to tune $T$. As observed from the figure, 
DP-FedAvg outperforms ISRL-DP (due to its use of local updates). DP-SCAFFOLD achieves better results than DP-FedAvg by correcting for client drift. 
Our proposed algorithm surpasses DP-SCAFFOLD, because of the dynamic allocation strategy that maintains a stable performance also for very large values of $T$.

\subsection{Nonsmooth Problem}
In this set of experiments, we consider the  
nonsmooth logistic regression problem
	\begin{align}
		\hspace{-2mm}\underset{{x} \in \mathbb{R}^{d}}{\operatorname{min}} \;\frac{1}{n} \sum_{i=1}^n f_i(x)+\frac{\vartheta_2}{2}\|{x}\|^2+ g(x), 
	\end{align}
	where $g(x)$ is the weighted $\ell_1$-norm over a box defined in Section~\ref{sec:analysis} with $\omega_j=0.01$ for all $j\in [d]$ and $\alpha=10$. 
Compared to problem  \eqref{eq-logreg}, we add a nonsmooth regularizer $g$, while all the other settings are the same. 
We use \texttt{a6a} dataset and set $n=20$, $m=561$, and $\vartheta_2=0.1$.  
\vspace{-1mm}
\begin{figure}[htbp]
	\centering	\includegraphics[width=5.6cm]{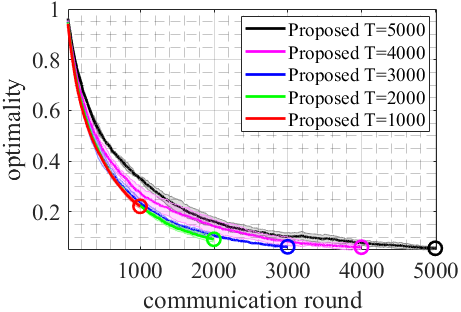}
	\caption{Nonsmooth problem on a6a satisfying $(1,10^{-4})$-LDP.}
	\label{fig-T}
\end{figure}
The aim of these experiments is to verify that our algorithm can run for a large but finite $T$  while ensuring $(1,10^{-4})$-LDP also on nonsmooth problems. To do so, we set different values of $T$ ranging from 1000 to 5000, and show the evolution of the optimality error across communication rounds $t=1, 2, \dots, T$. As shown in Fig.~\ref{fig-T},  our algorithm is effective for a range of $T$ values. As $T$ increases from 1000 to 3000, the optimality error becomes better. As $T$ increases from 3000 to 5000, the optimality error remains similar. This confirms that, in contrast to the prior state-of-the-art, our algorithm does not diverge even if we use an excessively large value of $T$.

\vspace{-2mm}
\section{Conclusions}
We have proposed an LDP FL algorithm for strongly convex, possibly nonsmooth, problems that protects the gradient of each worker against an honest but curious server. The dynamic allocation of the noise variance, which is derived by minimizing the optimization error subject to a privacy budget constraint,  removes the need for tuning the total number of communication rounds. Our algorithm outperforms state-of-the-art methods, making it a promising approach for privacy-preserving FL. Future research will explore data subsampling amplification and extensions to decentralized architectures.

\vspace{-1mm}

\begin{thebibliography}{10}

\bibitem{li2020federated}
Tian Li, Anit~Kumar Sahu, Ameet Talwalkar, and Virginia Smith,
\newblock ``Federated learning: Challenges, methods, and future directions,''
\newblock {\em IEEE Signal Processing Magazine}, vol. 37, no. 3, pp. 50--60,
  2020.

\bibitem{zhu2019deep}
Ligeng Zhu, Zhijian Liu, and Song Han,
\newblock ``Deep leakage from gradients,''
\newblock {\em Advances in Neural Information Processing Systems}, vol. 32,
  2019.

\bibitem{chou2020privacy}
Edward~J Chou, Arun Gururajan, Kim Laine, Nitin~Kumar Goel, Anna Bertiger, and
  Jack~W Stokes,
\newblock ``Privacy-preserving phishing web page classification via fully
  homomorphic encryption,''
\newblock in {\em IEEE International Conference on Acoustics,
  Speech and Signal Processing}, 2020, pp. 2792--2796.

\bibitem{dwork2006calibrating}
Cynthia Dwork, Frank McSherry, Kobbi Nissim, and Adam Smith,
\newblock ``Calibrating noise to sensitivity in private data analysis,''
\newblock in {\em Theory of Cryptography: Third Theory of Cryptography
  Conference}, 2006, pp. 265--284.

\bibitem{dwork2008differential}
Cynthia Dwork,
\newblock ``Differential privacy: A survey of results,''
\newblock in {\em International Conference on Theory and Applications of Models
  of Computation}, 2008, pp. 1--19.

\bibitem{hu2021concentrated}
Rui Hu, Yuanxiong Guo, and Yanmin Gong,
\newblock ``Concentrated differentially private federated learning with
  performance analysis,''
\newblock {\em IEEE Open Journal of the Computer Society}, vol. 2, pp.
  276--289, 2021.


\bibitem{zhao2022fldp}
Dan Zhao, Hong Chen, Suyun Zhao, Ruixuan Liu, Cuiping Li, and Xiaoying Zhang,
\newblock ``FLDP: Flexible strategy for local differential privacy,''
\newblock in {\em IEEE International Conference on Acoustics,
  Speech and Signal Processing}, 2022, pp. 2974--2978.


\bibitem{noble2022differentially}
Maxence Noble, Aur{\'e}lien Bellet, and Aymeric Dieuleveut,
\newblock ``Differentially private federated learning on heterogeneous data,''
\newblock in {\em International Conference on Artificial Intelligence and
  Statistics}, 2022, pp. 10110--10145.

\bibitem{lowy2022private}
A.~Lowy and M.~Razaviyayn, ``Private federated learning without a trusted
  server: Optimal algorithms for convex losses,'' in \emph{The Eleventh
  International Conference on Learning Representations}, 2022.

\bibitem{karimireddy2020scaffold}
Sai~Praneeth Karimireddy, Satyen Kale, Mehryar Mohri, Sashank Reddi, Sebastian
  Stich, and Ananda~Theertha Suresh,
\newblock ``Scaffold: Stochastic controlled averaging for federated learning,''
\newblock in {\em International Conference on Machine Learning}, 2020, pp.
  5132--5143.

\bibitem{mcmahan2018learning}
H.~B. McMahan, D.~Ramage, K.~Talwar, and L.~Zhang, ``Learning differentially
  private recurrent language models,'' in \emph{International Conference on
  Learning Representations}, 2018.

\bibitem{bun2016concentrated}
Mark Bun and Thomas Steinke,
\newblock ``Concentrated differential privacy: Simplifications, extensions, and
  lower bounds,''
\newblock in {\em Theory of Cryptography Conference}. Springer, 2016, pp.
  635--658.

\bibitem{alghunaim2019linearly}
Sulaiman Alghunaim, Kun Yuan, and Ali~H Sayed,
\newblock ``A linearly convergent proximal gradient algorithm for decentralized
  optimization,''
\newblock {\em Advances in Neural Information Processing Systems}, vol. 32,
  2019.

\bibitem{mishchenko2022proxskip}
Konstantin Mishchenko, Grigory Malinovsky, Sebastian Stich, and Peter
  Richt{\'a}rik,
\newblock ``Proxskip: Yes! local gradient steps provably lead to communication
  acceleration! finally!,''
\newblock in {\em International Conference on Machine Learning}, 2022, pp.
  15750--15769.

\bibitem{Bertsekas2003}
Dimitri Bertsekas,
\newblock {\em Nonlinear Programming},
\newblock Athena Scientific, 2003.

\bibitem{beck2017first}
Amir Beck,
\newblock {\em First-order methods in optimization},
\newblock SIAM, 2017.
\end{thebibliography}

\begin{thebibliography}{10}
	
	\bibitem[A1]{renyi1961measures}
	Alfr{\'e}d R{\'e}nyi,
	\newblock ``On measures of entropy and information,''
	\newblock in {\em Proceedings of the Fourth Berkeley Symposium on Mathematical
		Statistics and Probability, Volume 1: Contributions to the Theory of
		Statistics}. University of California Press, 1961, vol.~4, pp. 547--562.
	
	\bibitem[A2]{nesterov2003introductory}
	Yurii Nesterov,
	\newblock {\em Introductory lectures on convex optimization: A basic course},
	vol.~87,
	\newblock Springer Science \& Business Media, 2003.
	
\end{thebibliography}

\newpage
\appendix
\section{Appendix}\label{section-app}
\subsection{Proof of Lemma \ref{lem-total-dp}}\label{app-proof-total-dp}

Before proving the privacy loss, we give the definition of $\rho$-zCDP,  two useful facts on the composition property of $\rho$-zCDP \cite{dwork2006calibrating},  and the equivalent relation of $\rho$-zCDP and $(\epsilon, \delta)$-DP \cite{bun2016concentrated}. 

\begin{definition}[$\rho$-zCDP] 
	\label{def-zcdp}
	A randomized algorithm $\mathcal{A}$ is said to satisfy $\rho$-zCDP if, for all $\alpha \in(1, \infty)$ and all adjacent inputs $\nabla f_i, \nabla f'_i$, the following holds:
	\begin{equation}
		D_{\alpha}\left(\mathcal{A}(\nabla f_i) \| \mathcal{A}(\nabla f_i')\right)
		\leq \alpha \rho,
	\end{equation} 
	where $D_\alpha$ is the Rényi divergence \cite{renyi1961measures}.
\end{definition}

\begin{fact}[Gaussian mechanism] \label{fct-gaussian-mechanism}
	Let $\mathcal{Q}$ be a function with sensitivity $\Delta_\mathcal{Q}$. Then, the mechanism $\mathcal{A}(x) \sim \mathcal{N}(\mathcal{Q}(x), \xi^2 I_d)$ satisfies $(\Delta_\mathcal{Q}^2 / (2\xi^2))$-zCDP.
\end{fact}

\begin{fact}[Composition]
	\label{fct-composition}	
	Let  $\mathcal{A}(x)$$ = (\mathcal{A}_1(x), \ldots , \mathcal{A}_T(x))$ be an iterative randomized algorithm. If each $\mathcal{A}_t$ respectively satisfies $\rho_t$-zCDP, then running $\mathcal{A}(x)$
	satisfies $\left(\sum_{t=1}^T \rho_t\right)$-zCDP.
\end{fact}

\begin{fact}
	\label{fct-cdp-dp}	
	If a randomized algorithm $\mathcal{A}$ satisfies $\rho$-zCDP, then $\mathcal{A}$ satisfies $\left(\rho+2 \sqrt{\rho \ln {1}/{\delta}}, \delta\right)$-DP for any $\delta >0$.
\end{fact}

To bound the privacy loss, we will show that $\widetilde{x}_i^{t+1}$ is the output of a Gaussian mechanism with appropriately chosen variance and sensitivity.
Recall that $\widetilde{x}_i^{t+1} = {x}_i^t-\gamma\tfrac{1}{n}\nabla f_i({x}_i^t)-\gamma {\Lambda}_i^t-\gamma \zeta_{i}^t$.
Note that $x_i^t$ and $\Lambda_i^t$ are a data-independent processing of $\widetilde{x}_i^{t}$.
Define auxiliary functions $\mathcal{Q}_1(\widetilde{x}_i^{t}) = x_i^t - \gamma \Lambda_i^t$ and $\mathcal{Q}_2(\nabla f_i ;\, \widetilde{x}_i^{t}) = - \gamma \frac{1}{n} \nabla f_i(x_i^t)$. Then, $\widetilde{x}_i^{t+1}$ is a Gaussian mechanism for the query function $\mathcal{Q}(\nabla f_i;\, \widetilde{x}_i^{t}) = \mathcal{Q}_1(\widetilde{x}_i^{t}) + \mathcal{Q}_2(\nabla f_i;\, \widetilde{x}_i^{t})$. The sensitivity of $\mathcal{Q}$ is
\begin{equation}\label{eq-sens-alg}
	\begin{aligned}
		\Delta_{\mathcal{Q}}
		&= \max_{\nabla f_i, \nabla f'_i} \left\|\mathcal{Q}(\nabla f_i;\, \widetilde{x}_i^{t})-\mathcal{Q}(\nabla f'_i;\, \widetilde{x}_i^{t})\right\| \\
		&= \max_{\nabla f_i, \nabla f'_i} \left\|\mathcal{Q}_2(\nabla f_i;\, \widetilde{x}_i^{t})-\mathcal{Q}_2(\nabla f'_i;\, \widetilde{x}_i^{t})\right\| \\
		&= \max_{\nabla f_i, \nabla f'_i} \left\| \frac{\gamma}{nm}(\nabla f_{i\tilde{l}}-{\nabla f'_{i\tilde{l}}}) \right\| \leq \frac{2\gamma B}{nm},
	\end{aligned}
\end{equation}
where we assume (WLOG) the $\tilde{l}$-th gradient is different and we use Assumption \ref{asm:g-prox} to get $\|\nabla f_{i\tilde{l}}-{\nabla f'_{i\tilde{l}}}\|\le 2B$.
Then, by Fact \ref{fct-gaussian-mechanism}, each $\widetilde{x}_i^{t}$ satisfies $\rho^t$-zCDP with $\rho^{t}=\frac{2B^2}{ n^2m^2\xi_t^2}$.
By the composition property in Fact \ref{fct-composition}, during the whole iterative procedure, each $\widetilde{x}_i^{t}$  satisfies $\left(\sum_{t=1}^T\frac{2B^2}{ n^2m^2\xi_t^2}\right)$-zCDP for all $i$.  By Fact \ref{fct-cdp-dp}, we know that  each $\widetilde{x}_i^{t}$  satisfies $\left(\epsilon, \delta\right)$-LDP with $\epsilon=\sum_{t=1}^T\frac{2B^2}{ n^2m^2\xi_t^2}+2 \sqrt{\left(\sum_{t=1}^T\frac{2B^2}{ n^2m^2\xi_t^2}\right) \ln \frac{1}{\delta}} $ for all $i$.  We complete the proof.

\subsection{Proof of Lemma \ref{lem-descent}}\label{app-prof-lem-descent}
To simplify the analysis, we rewrite the proposed algorithm \eqref{eq-alg-fl} as 
\begin{equation}\label{eq-alg-analysis}
	\left\{\begin{aligned}
		\hat{\mathbf{x}}^{t+1} & =\mathbf{x}^t-\gamma\left(\nabla f\left(\mathbf{x}^t\right)+\mathbf{L} \mathbf{y}^t\right), \\
		\widetilde{\mathbf{x}}^{t+1} & =\hat{\mathbf{x}}^{t+1}-\gamma \boldsymbol{\zeta}^t, \\
		\mathbf{y}^{t+1} & =\mathbf{y}^t+\mathbf{L} \widetilde{\mathbf{x}}^{t+1}, \\
		\mathbf{z}^{t+1} & =\widetilde{\mathbf{x}}^{t+1}-\gamma \mathbf{L}\left(\mathbf{y}^{t+1}-\mathbf{y}^t\right) =(\mathbf{I}-\gamma \mathbf{L}) \widetilde{\mathbf{x}}^{t+1}, \\
		\mathbf{x}^{t+1} & =\operatorname{prox}_{\frac{\gamma}{n} g}\left(\mathbf{z}^{t+1}\right),
	\end{aligned}\right.   
\end{equation}
where we introduce an auxiliary variable $\hat{\mathbf{x}}^{t+1}$ and use the fact that $\mathbf{L}^2=\mathbf{L}$ in the fourth equality. The following analysis is based on \eqref{eq-alg-analysis}. 
To establish the convergence, let us begin with the claim that in the noiseless setting where $\bzeta^t=0$ the following holds   
\begin{align}
	&\bz^{\star}=\bx^{\star}- \gamma \nabla  f(\bx^{\star})-\gamma \mathbf{L} \by^{\star}, \label{eq-fix-8}\\
	&\mathbf{L}\bz^{\star}= \mathbf{L}\bx^{\star}=0_{nd},\label{eq-fix-9}\\ 
	&\bx^{\star}=\prox_{\frac{\gamma}{n}g} (\bz^{\star}) \label{eq-fix-10}
\end{align}
for any $\gamma>0$, where $\bx^{\star}=[x^{\star};\ldots; x^{\star}]$ with $x^{\star}$ satisfying the  optimality condition of problem \eqref{eqn:basic_opt}, i.e.,  
\begin{equation}\label{eq-kkt}
	\begin{aligned}
		-\frac{1}{n}\sum_{i=1}^{n} \nabla f_i(x^{\star}) \in \partial g(x^{\star}). 
	\end{aligned}
\end{equation}
To show  \eqref{eq-fix-8}-\eqref{eq-fix-10}, let us  construct $(\bx^{\star},\by^{\star}, \bz^{\star})$ with the optimal solution $x^{\star}$ of the problem \eqref{eqn:basic_opt}. 	
First, we construct $z^{\star}$ with $x^{\star}$ by
\begin{equation}\label{eq-z*}
	\sum_{i=1}^{n}z^{\star}=\sum_{i=1}^{n} x^{\star}-\frac{\gamma }{n} \sum_{i=1}^{n} \nabla f_i(x^{\star}). 
\end{equation}
Since $x^{\star}$ is the optimal solution of the problem \eqref{eqn:basic_opt}, we know that 
\eqref{eq-kkt} holds. Combing \eqref{eq-kkt} and \eqref{eq-z*}, we know that 
\begin{equation}\label{eq-x*}
	z^{\star}-x^{\star}\in \tfrac{\gamma}{n}\partial g(x^{\star}). 
\end{equation}	
By setting $\bx^{\star}=[x^{\star};\ldots;x^{\star}]$ and $\bz^{\star}=[z^{\star};\ldots;z^{\star}]$, we know that such $\bx^{\star}$ and $\bz^{\star}$ satisfy both \eqref{eq-fix-9} and \eqref{eq-fix-10}. Next, we construct $\by^{\star}$. 
With \eqref{eq-x*}, we know that 
\begin{equation*}
	\begin{aligned}
		&\bx^{\star}-\gamma \nabla f(\bx^{\star})-\bz^{\star}\\
		\in& -\tfrac{\gamma}{n}\left[\partial g(x^{\star})+\nabla f_1(x^{\star});\ldots;\partial g(x^{\star})+\nabla f_n(x^{\star})\right], 
	\end{aligned}	 
\end{equation*}
where $\in$ is defined block-wisely. This implies that there exists $\by^{\star}$ satisfying the following condition 
\begin{equation}
	\bz^{\star}=\bx^{\star}- \gamma \nabla  f(\bx^{\star})-\gamma \mathbf{L} \by^{\star}. 
\end{equation}
So far, we have shown \eqref{eq-fix-8}-\eqref{eq-kkt}.
In the following, we  prove the main recursion of our Lyapunov function $\Phi^t$.    		
With the update of $\bz^{t+1}$ in \eqref{eq-alg-analysis}, we compute
\begin{equation}\label{eq-x1}
	\begin{aligned}
		&\left\|{\bz^{t+1}-\bz^{\star}}\right\|^{2}\\ =&\left\|\hat{\bx}^{t+1}-{\bz^{\star}}-\gamma \mathbf{L}\left(\by^{t+1}-\by^{t}\right) -{\gamma \bzeta^t}\right\|^{2}\\
		=&\left\|\hat{\bx}^{t+1}-{\bz^{\star}}\right\|^{2}-2{\gamma}\left\langle\hat{\bx}^{t+1}-{\bx^{\star}}, \mathbf{L}\left(\by^{t+1}-\by^{t}\right)\right\rangle\\
		&+{\gamma^{2}}\left\|\mathbf{L}\left(\by^{t+1}-\by^{t}\right) +{\bzeta^t}\right\|^{2} {-2{\gamma}\left\langle\hat{\bx}^{t+1}-{\bz^{\star}}, \bzeta^t\right\rangle},
	\end{aligned}
\end{equation}
where we use $\mathbf{L}\bz^{\star}=\mathbf{L}\bx^{\star}=0_{nd}$ in the last equality. 

Next, let us bound the first term $\left\|\hat{\bx}^{t+1}-{\bz^{\star}}\right\|$ on the right hand of \eqref{eq-x1}. To this end, we define $\bw^t=\bx^t-\gamma \nabla 
f(\bx^t)$ and  $\bw^{\star}=\bx^{\star}-\gamma \nabla 
f(\bx^{\star})$, we have  
\begin{equation}\label{eq-x2}
	\begin{aligned}
		&\left\|\hat{\bx}^{t+1}-\bz^{\star}\right\|^{2}\\
		=&\left\| \bx^t-\gamma\nabla f(\bx^t)-\gamma \mathbf{L}\by^t-\bz^{\star}\right\|^{2}\\
		=&\left\|\bw^{t}-\bw^{\star}-\gamma \mathbf{L}\left(\by^{t}-\by^{\star}\right)\right\|^{2}\\
		=&\left\|\bw^{t}-\bw^{\star}\right\|^{2}-2 \gamma\left\langle\hat{\bx}^{t+1}-{\bx^{\star}}, \mathbf{L}\left(\by^{t}-\by^{\star}\right)\right\rangle\\
		&-\gamma^{2}\left\|\mathbf{L}\left(\by^{t}-\by^{\star}\right)\right\|^{2},
	\end{aligned}
\end{equation}
where we use {$\bz^{\star}=\bx^{\star}-\gamma \nabla f(\bx^{\star})-\gamma \mathbf{L}\by^{\star}=\bw^{\star}-\gamma \mathbf{L}\by^{\star}$} in the second equality and  $\|a+b\|^{2}=\|a\|^{2}+2\langle a+b, b\rangle-\|b\|^{2}$ and $\mathbf{L}\bx^{\star}=\mathbf{L}\bz^{\star}=0_{nd}$ in the last equality. 

Thus, by substituting \eqref{eq-x2} into \eqref{eq-x1}, we have 
\begin{equation}\label{eq-x3}
	\begin{aligned}
		&\left\|{\bz^{t+1}}-\bz^{\star}\right\|^{2}\\
		=&\left\|\bw^{t}-\bw^{\star}\right\|^{2}-2 \gamma\left\langle\hat{\bx}^{t+1}-\bx^{\star}, \mathbf{L}\left(\by^{t}-\by^{\star}\right)\right\rangle\\
		-&2 \gamma\left\langle\hat{\bx}^{t+1}-{\bx^{\star}}, \mathbf{L}\left(\by^{t+1}\!-\!\by^{t}  \right)\right\rangle\!-\!\gamma^{2}\left\|\mathbf{L}\left(\by^{t}\!-\!\by^{\star}\right)\right\|^{2}\\
		+&\gamma^{2}\left\|\mathbf{L}\left(\by^{t+1}-\by^{t}\right){+\bzeta^t}\right\|^{2} {-2{\gamma}\left\langle\hat{\bx}^{t+1}-{\bz^{\star}}, \bzeta^t\right\rangle}\\
		=&\left\|\bw^{t}-\bw^{\star}\right\|^{2}-2 \gamma\left\langle\hat{\bx}^{t+1}-\bx^{\star}, \mathbf{L}\left(\by^{t+1}-\by^{\star}\right)\right\rangle\\
		&-\gamma^{2}\left\|\mathbf{L}\left(\by^{t}-\by^{\star}\right)\right\|^{2}+\gamma^{2}\left\|\mathbf{L}\left(\by^{t+1}-\by^{t}\right) { + \bzeta^t } \right\|^{2}\\
		&{-2{\gamma}\left\langle\hat{\bx}^{t+1}-{\bz^{\star}}, \bzeta^t\right\rangle}.
	\end{aligned}
\end{equation}
Now, we start to bound  $\|\by^{t+1}-\by^{\star}\|^2$.  According to \eqref{eq-alg-analysis}, we have
\begin{equation}\label{eq-y-1}
	\begin{aligned}
		&\left\|\by^{t+1}-\by^{\star}\right\|^{2}\\ =&\left\|\by^{t}-\by^{\star}+\left(\by^{t+1}-\by^{t}\right)\right\|^{2} \\
		=&\left\|\by^{t}-\by^{\star}\right\|^{2}+2 \left\langle \by^{t+1}-\by^{\star}, \by^{t+1}-\by^{t}\right\rangle-\left\|\by^{t+1}-\by^{t}\right\|^{2} \\
		=&\left\|\by^{t}-\by^{\star}\right\|^{2}+2 \left\langle \by^{t+1}-\by^{\star}, \mathbf{L} (\hat{\bx}^{t+1}-{\gamma\bzeta^t})\right\rangle\\
		&-\left\|\by^{t+1}-\by^{t}\right\|^{2}.
	\end{aligned}
\end{equation}
Combining \eqref{eq-x3} and \eqref{eq-y-1}, we have 
\begin{equation}\label{eq-lyapunov-1}
	\begin{aligned}
		&\left\|{\bx^{t+1}}-\bx^{\star}\right\|^{2}+{\gamma}\left\|\by^{t+1}-\by^{\star}\right\|^{2}	\\
		\le&\left\|{\bz^{t+1}}-\bz^{\star}\right\|^{2}+{\gamma}\left\|\by^{t+1}-\by^{\star}\right\|^{2} \\
		\le &\left\|\bw^{t}-\bw^{\star}\right\|^{2}+{\gamma}\left\|\by^{t}-\by^{\star}\right\|^{2}\!-{\!\gamma^{2}\left\|\mathbf{L}\left(\by^{t}-\by^{\star}\right)\right\|^{2}}\\
		&-{\gamma}\left\|\by^{t+1}-\by^{t}\right\|^{2}+\gamma^{2}\left\|\mathbf{L}\left(\by^{t+1}-\by^{t}\right) { + \bzeta^t} \right\|^{2}\\
		&-2\gamma \left\langle \by^{t+1}-\by^{\star}, \mathbf{L} {\gamma\bzeta^t}\right\rangle{-2{\gamma}\left\langle\hat{\bx}^{t+1}-{\bz^{\star}}, \bzeta^t\right\rangle},
	\end{aligned}
\end{equation}
where we use $\|\prox_{\frac{\gamma}{n}g} (\bz^{t+1}) -\prox_{\frac{\gamma}{n}g}(\bz^{\star})\|\le \left\|{\bz^{t+1}}-\bz^{\star}\right\| $ in the first inequality. 
Next, we bound the first term $\left\|\bw^{t}-\bw^{\star}\right\|^{2}$ on the right hand of \eqref{eq-lyapunov-1} as 
\begin{equation}\label{eq-w-linear}
	\begin{aligned}
		&\left\|\bw^{t}-\bw^{\star}\right\|^{2}\\
		=& \left\|\bx^t-\bx^{\star}-\gamma \nabla f(\bx^t)+\gamma \nabla f\left(\bx^{\star}\right)\right\|^2 \\
		=&\left\|\bx^t-\bx^{\star}\right\|^2-2 \gamma\left\langle\bx^t-\bx^{\star}, \nabla f\left(\bx^t\right)-\nabla f\left(\bx^{\star}\right)\right\rangle\\
		&+\gamma^2 \left\|\nabla f\left(\bx^t\right)-\nabla f\left(\bx^{\star}\right)\right\|^2 \\
		\leq &(1-\gamma \mu_f)\left\|\bx^t-\bx^{\star}\right\|^2,
	\end{aligned}
\end{equation}
where we use 
$
\| \bx^{t}-\gamma \nabla f(\bx^t) -\bx^{\star}+\gamma \nabla f(\bx^{\star}) \|^2\le (1-\gamma\mu_f)\| \bx^t-\bx^{\star}\|^2
$ 
when $\gamma \le {1}/{L_f}$  in the last inequality  \cite[Theorem 2.1.15]{nesterov2003introductory}. 

Thus, by substituting \eqref{eq-w-linear} into \eqref{eq-lyapunov-1},  we have 
\begin{equation}\label{eq-linear-inner-product}
	\begin{aligned}
		&\mathbb{E} \left[ \left\|{\bx^{t+1}}-\bx^{\star}\right\|^{2}+{\gamma}\left\|\by^{t+1}-\by^{\star}\right\|^{2}\mid\mcf^t\right]\\
		\le&{(1-\gamma\mu_f)}\|{\bx^t}-\bx^{\star}\|^2+ {\left( 1-\gamma \right){\gamma}\left\|\by^{t}-\by^{\star}\right\|^{2}}\\
		-&{\gamma}\bE\left[  \left\|\by^{t+1}-\by^{t}\right\|^{2}\mid \mcf^t\right]-2\gamma\bE \left[ \left\langle \by^{t+1}-\by^{\star}, \mathbf{L} {\gamma\bzeta^t}\right\rangle \mid \mathcal{F}^t \right]\\
		+&\gamma^{2} \bE\left[ \left\|\mathbf{L}\left(\by^{t+1}-\by^{t}\right) {+ \bzeta^t}\right\|^{2}\mid \mcf^t\right],
	\end{aligned}
\end{equation}
where we use  $\|\mathbf{L} (\by^t-\by^*) \|^2 \ge \lambda_{\min}^+ (\mathbf{L}^2) \| \by^t-\by^* \|^2 \ge \| \by^t-\by^* \|^2$ which is given by the facts that $\mathbf{L}=(I_n-1_n 1_n^T/n)\otimes I_d $, $\mathbf{L}^2=\mathbf{ L}$, and $\lambda_{\min}^+(\mathbf{L}) =1$ where $\lambda_{\min}^+(\cdot)$ denotes the smallest positive eigenvalue. 
Next, we bound the last term on the right hand of \eqref{eq-linear-inner-product} as 
\begin{equation}\label{eq-16}
	\begin{aligned}
		-&2 \gamma^2 \bE\left[  \left\langle \by^{t+1}-\by^{\star}, \mathbf{L} {\bzeta^t}\right\rangle \mid \mcf^t\right]\\
		=&2\gamma^2 \left( \bE\left[ \left\langle \by^{t+1}\!-\!\by^{t},  \!-\mathbf{L}{\bzeta^t}\right\rangle \mid \mcf^t\right]\!-\!\bE\left[ \left\langle  \by^{t}\!-\!\by^{\star}, \mathbf{L} {\bzeta^t}\right\rangle \mid \mcf^t\right]  \right)   \\
		=&2\gamma^2 \bE\left[ \left\langle  \by^{t+1}-\by^{t}, -\mathbf{L} {\bzeta^t}\right\rangle \mid \mcf^t\right]  \\
		\le& \frac{\gamma}{2} \bE[ \|\by^{t+1}-\by^{t} \|^2\mid \mcf^t] +2\gamma^3nd \xi_t^2, 
	\end{aligned}
\end{equation}
where we use $\bE [\bzeta^t|\mathcal{F}^t]=0$ in the second equality and  $ 2\langle a,b \rangle\le \|a\|^2+\|b\|^2 $ in the last inequality. 
Substituting \eqref{eq-16} into \eqref{eq-linear-inner-product}, we get 
\begin{equation}\label{eq-17}
	\begin{aligned}
		&\bE\left[\Phi^{t+1}\mid \mcf^t\right]\\
		\le& \max \left\{ 1-\gamma\mu_f,  1-\gamma \right\} \Phi^t+2{\gamma^3} nd \xi_t^2\\
		-&{\frac{\gamma}{2}}\bE\left[  \left\|\by^{t+1}\!-\!\by^{t}\right\|^{2}\mid \mcf^t\right]\!+\!\gamma^{2} \bE\left[ \left\|\mathbf{L}\left(\by^{t+1}\!-\!\by^{t}\right) {+ \bzeta^t}\right\|^{2}\mid \mcf^t\right]\\
		\le& \max \left\{ 1-\gamma\mu_f,  1-\gamma \right\} \Phi^t+2{\gamma^3} nd \xi_t^2\\
		&+{\left(-\frac{\gamma}{2} +2\gamma^2 \right)\bE\left[  \left\|\by^{t+1}-\by^{t}\right\|^{2}\mid \mcf^t\right] + 2\gamma^2 nd\xi_t^2  }\\
		\le&  \max \left\{ 1-\gamma\mu_f,  1-\gamma \right\} \Phi^t+2(1+\gamma){\gamma^2} nd \xi_t^2,
	\end{aligned}
\end{equation}
where we use $\lambda_{\max}(\mathbf{L} )=1$ and $\gamma\le 1/4$. Thus, we get \eqref{eq-one-descent} and complete the proof. 

\subsection{Proof of Theorem \ref{thm-final}}\label{app-prof-thm-final} 
Telescoping \eqref{eq-one-descent} from $t=1$ to $t=T$, we get 
\begin{equation}\label{eq-40}
	\begin{aligned}
		&\bE\left[\Phi^{T+1} \right]\\
		\le& (1-\gamma C_1)^T \bE\left[\Phi^{1} \right] + C_2 \gamma^2 \left( q_1 \xi_1^2 +\ldots+ q_T \xi_T^2 \right),
	\end{aligned}
\end{equation}
where we define $q_t=(1-\gamma C_1)^{T-t}$,  $C_1=\min\left\{{\mu_f},1\right\}$ and $C_2=2.5nd$. 
In order to dynamically allocate the privacy budget, we  solve the following optimization problem 
\begin{equation}\label{eq-dynm-p}
	\begin{aligned}
		\operatorname*{minimize}_{\{\xi_t^2\}_{t=1}^{T}}~ &q_1 \xi_1^2 +\cdots+ q_T\xi_T^2,\\
		\text{s.t.}~& \frac{2B^2}{n^2{m^2}}\left(\frac{1}{\xi_1^2}+\cdots +\frac{1}{ \xi_T^2}\right) =\rho_{\rm tgt},
	\end{aligned}
\end{equation}
which can be verified to be strictly convex via reformulation. To solve the problem \eqref{eq-dynm-p}, we write down its KKT condition 
\begin{equation}\label{eq-kkt-dynm}
	q_t -\frac{\pi}{(\xi_t^2)^2}=0, \quad \frac{1}{\xi_1^2}+\cdots +\frac{1}{ \xi_T^2} = \frac{\rho_{\rm tgt} n^2{m^2}}{2B^2}, 
\end{equation}
where $\pi\in\mathbb{R}$ is the dual variable. 
From \eqref{eq-kkt-dynm}, we get the solution of problem \eqref{eq-dynm-p} 
\begin{equation}\label{eq-dynm-soln}
	\xi_t^2= \sqrt{\frac{\pi}{q_t} }, \quad \sqrt{\pi}= \frac{\sum_{t=1}^T\sqrt{q_t}}{\frac{\rho_{\rm tgt} n^2{m^2}}{2B^2}}, \quad \forall t\in [T].  
\end{equation}
Substituting the solution \eqref{eq-dynm-soln} into \eqref{eq-40}, we get 
\begin{equation}\label{eq-44}
	\begin{aligned}
		&\bE\left[ \Phi^{T+1} \right]\\
		\le& (1-\gamma C_1)^T \bE\left[ \Phi^{1} \right] + \frac{C_2 \gamma^22B^2}{{\rho_{\rm tgt} n^2{m^2}}} \left(\sum_{p=0}^{T-1}
		\left(\sqrt{1-\gamma C_1}\right)^p \right)^2\\
		\le &(1-\gamma C_1)^T \bE\left[ \Phi^{1} \right]+ \frac{C_2 \gamma^22B^2}{{\rho_{\rm tgt} n^2{m^2}}} {\cdot \frac{1}{(1-\sqrt{1-\gamma C_1})^2}}\\
		= & (1-\gamma C_1)^T \bE\left[ \Phi^{1} \right]+ \frac{C_22B^2 }{\rho_{\rm tgt} n^2{m^2}} \frac{ (1+\sqrt{1-\gamma C_1})^2 }{C_1^2} \\
		\le&(1-\gamma \min\left\{{\mu_f},1\right\}   )^T \bE\left[ \Phi^{1} \right]\\
		& + \frac{20  B^2d }{ \left(\sqrt{\epsilon+\ln \tfrac{1}{\delta}}-\sqrt{\ln \tfrac{1}{\delta}}\right)^2 n{m^2}\min\left\{\mu_f^2,1\right\}}, 
	\end{aligned}
\end{equation}
where we use the fact that 
$
\rho_{\rm tgt}= \left(\sqrt{\epsilon+\ln {1}/{\delta} }-\sqrt{\ln {1}/{\delta}}\right)^2
$
and  substitute $C_1=\min\left\{{\mu_f},1\right\}$ and $C_2=2.5nd$. If we substitute 
\begin{equation*}
	\left(\sqrt{\epsilon+\ln \frac{1}{\delta} }-\sqrt{\ln \frac{1}{\delta}}\right)^2 \approx \frac{\epsilon^2}{4\ln \frac{1}{\delta}},
\end{equation*}
\eqref{eq-44} implies that 
\begin{equation*}
\begin{aligned}
  \bE\left[ \Phi^{T+1} \right]\le&  (1\!-
	\!\gamma \min\left\{{\mu_f},1\right\}   )^T \bE\left[ \Phi^{1} \right]\\
 &+\mathcal{O}\left( \!\frac{B^2 d \ln \frac{1}{\delta}}{\epsilon^2nm^2\min\left\{{\mu_f^2},1\right\} } \!\right).   
\end{aligned}	
\end{equation*}
As $(1-z)^T\leq e^{-Tz}$, if $ T\ge  \frac{\ln \left(\frac{\epsilon^2nm^2\min\left\{{\mu_f^2},1\right\} \bE\left[ \Phi^{1} \right]}{B^2 d \ln \frac{1}{\delta}} \right)}{\gamma \min\{\mu_f,1\}} $, then
\begin{equation}
	\begin{aligned}
		\bE\left[ \Phi^{T+1} \right]
		=\mathcal{O}\left( \frac{B^2 d \ln \frac{1}{\delta}}{\epsilon^2nm^2\min\left\{{\mu_f^2},1\right\} } \right).
	\end{aligned}		
\end{equation}
This completes the proof.

\subsection{Additional Experiments: Classification Using Neural Networks.}
In this set of experiments, we solve a nonconvex nonsmooth problem that involves classifying the MNIST dataset \cite{noble2022differentially} using a neural network.  
It should be noted that our theoretical findings are no longer applicable to nonconvex problems. Here is a heuristic extension of the  proposed algorithm. 
We utilize a one-hidden-layer neural network (NN) to classify the MNIST dataset \cite{noble2022differentially}. 
The NN consists of 200 neurons in the hidden layer, employing a sigmoid activation function, and a softmax activation in the output layer. 
The loss function employed is cross-entropy with a regularization term $g(x)$ that is the weighted $\ell_1$-norm over a box defined in Section \ref{sec:analysis} with $\omega_j=10^{-5}$ for all $j\in [d]$ and $\alpha=10$. 
The MNIST dataset comprises grayscale images of handwritten digits, each with dimensions $28\times28$ pixels, and includes 10 classes representing digits from 0 to 9. 
Our training dataset consists of 38,500 samples. 
Noted that our algorithm and theory support heterogeneous data, we uniformly sample 10,000 samples from the entire training dataset, allocating 1,000 samples to each of the 10 nodes. The remaining 28,500 samples are assigned labels ranging from 0 to 9, with each class having 2,850 samples, distributed across nodes 1 to 10. We use 10,000 samples for testing.
We define two sets of privacy budgets as $(1, 10^{-4})$-LDP and $(0.5, 10^{-5})$-LDP, as well as three sets of total communication rounds as 500, 1000, and 1200.  
\begin{figure}[htbp]
	\begin{minipage}[h]{.48\linewidth}
		\centering
		\centerline{\includegraphics[width=4.3cm]{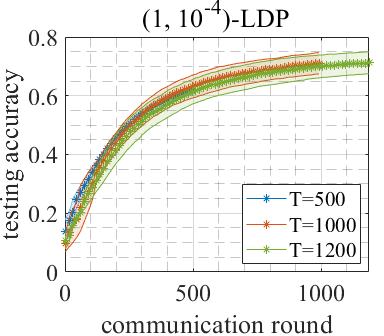}}
	\end{minipage}
	\hfill
	\begin{minipage}[h]{0.48\linewidth}
		\centering
		\centerline{\includegraphics[width=4.3cm]{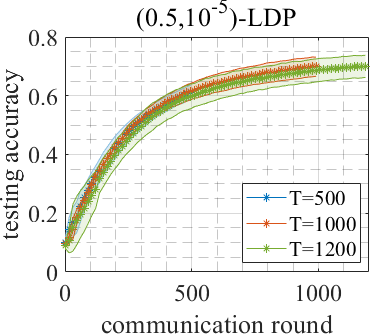}}
	\end{minipage}
	\caption{Classification of MNIST using neural networks.}
	\label{fig_dnn}
\end{figure}
Fig.~\ref{fig_dnn} shows the testing accuracy with respect to communication rounds.  
It is noted that insufficient communication rounds (500 rounds) result in lower accuracy. However, increasing the communication rounds to 1000 leads to an accuracy of around 70\%. Despite  more privacy-preserving noise being added with 1200 communication rounds, our algorithm consistently stabilizes at around 70\% accuracy. This means that practical applications can benefit from estimating an appropriate number of communication rounds, but even an overestimation does not compromise accuracy.

\end{document}